\documentclass{article}
\usepackage{spconf,amsmath,graphicx}
\usepackage{booktabs}
\usepackage[dvipsnames, table]{xcolor}
\usepackage{multirow}
\usepackage{url}
\usepackage{subcaption}
\usepackage{pifont}
\usepackage[backend=bibtex,citestyle=numeric-comp,bibstyle=ieee,defernumbers=true,giveninits=true,doi=false,isbn=false,url=false,eprint=false,minbibnames=10,maxbibnames=10]{biblatex}
\AtEveryBibitem{
}

\newcommand{\cmark}{\ding{51}}
\newcommand{\xmark}{\ding{55}}
\definecolor{LightCyan}{rgb}{0.88,1,1}

\ninept

\title{BRAVEn: Improving Self-Supervised Pre-training for Visual and Auditory Speech Recognition}
\name{Alexandros Haliassos\textsuperscript{*}\thanks{\textsuperscript{*}Equal contribution}, Andreas Zinonos\textsuperscript{*}, Rodrigo Mira, Stavros Petridis, Maja Pantic}
\address{Imperial College London}
\begin{document}
\maketitle
\begin{abstract}
Self-supervision has recently shown great promise for learning visual and auditory speech representations from unlabelled data. In this work, we propose BRAVEn, an extension to the recent RAVEn method, which learns speech representations entirely from raw audio-visual data. Our modifications to RAVEn enable BRAVEn to achieve state-of-the-art results among self-supervised methods in various settings. Moreover, we observe favourable scaling behaviour by increasing the amount of unlabelled data well beyond other self-supervised works. In particular, we achieve 20.0\,\% / 1.7\,\% word error rate for VSR / ASR on the LRS3 test set, with only 30 hours of labelled data and no external ASR models. Our results suggest that readily available unlabelled audio-visual data can largely replace costly transcribed data. Code at \url{https://github.com/ahaliassos/raven}.
\end{abstract}
\section{Introduction}
Visual and auditory speech recognition (VSR / ASR), \textit{i.e.}, the prediction of spoken words from visual and auditory inputs respectively, scale very well with the amount of transcribed data~\cite{makino2019recurrent, afouras2021sub, serdyuk2021audio, DBLP:journals/corr/abs-2201-10439}. However, the cost of accurately annotating large-scale datasets can be prohibitive; indeed, the largest publicly available labelled audio-visual dataset has under 500 hours of footage~\cite{afouras2018lrs3}. This issue has spurred significant research into bypassing annotated data requirements. 

One line of work along this direction is to use off-the-shelf ASR models to transcribe visual data (lip movement videos) and then train VSR models with a mix of labelled and pseudo-labelled data~\cite{ma2022visual,ma2023auto}. Similarly,~\cite{afouras2020asr} uses the outputs of an ASR model for cross-modal distillation. A different approach is to synthesise lip movements using transcribed audio, which in general is more abundant than transcribed visual data~\cite{liu2023synthvsr}. Although such approaches obtain impressive results, they still assume access to large-scale transcribed speech data, either to train the ASR model used for pseudo-labelling / distillation, or to assign labels to the corresponding synthesised visual data. It has also been observed that supervised training of VSR models may suffer from optimisation difficulties and may thus require intricate training strategies such as curriculum learning~\cite{ma2022visual,haliassos2022jointly}.

Another line of research, free from these drawbacks, is audio-visual self-supervised learning. In this framework, representations are first learned by exploiting the fine-grained correspondence between the visual and auditory modalities, and then the resulting encoders are fine-tuned on potentially less labelled data for speech recognition~\cite{ma2021lira,shi2022learning,haliassos2022jointly,zhu2023vatlm,baevski2022data2vec}. For example, LiRA~\cite{ma2021lira} learns a visual encoder by predicting audio PASE+ features~\cite{ravanelli2020multi}. In contrast, AV-HuBERT~\cite{shi2022learning} learns audio-visual features by predicting clustering indices generated in multiple stages, initially using MFCC features. VATLM~\cite{zhu2023vatlm} extends AV-HuBERT by incorporating text data into the pre-training process. Recently, student-teacher frameworks have been proposed~\cite{haliassos2022jointly,lian2023av} which, given masked inputs, predict contextualised targets generated by momentum teachers that are presented with unmasked inputs. In particular, RAVEn~\cite{haliassos2022jointly} uses separate video and audio students which regress the outputs of the teacher networks through lightweight Transformer predictors, and learns strong visual and auditory speech representations entirely from raw data.

In this work, we extend RAVEn (\textbf{R}aw \textbf{A}udio-\textbf{V}isual \textbf{En}coders) through modifications that yield better representations by acknowledging the semantic asymmetries between audio and video. We dub our approach Better RAVEn, or BRAVEn. Our enhancements are as follows: (1) We use the average of the outputs of each Transformer encoder block as our targets~\cite{lian2023av}, rather than the output of the last block, in order to create smoother targets; (2) We use a shallower predictor for the video student, which encourages the video encoder to better capture the information embedded in the predicted audio targets; (3) We use stronger masking for the audio inputs to address the difference in relative difficulty between VSR and ASR; (4) Finally, we use different loss weights for the audio predictors, which empirically benefits ASR performance. 

We find that BRAVEn scales well both with the model size as well as the amount of unlabelled data, consistently achieving state-of-the-art performance across self-supervised methods in comparable settings. We also explore increasing the quantity of unlabelled data beyond what has been used in competing methods by combining LRS3~\cite{afouras2018lrs3}, Voxceleb2~\cite{DBLP:conf/interspeech/ChungNZ18}, and AVSpeech~\cite{ephrat2018looking}. We observe that, keeping the amount of labelled data fixed, this leads to further significant improvements, especially for VSR. Notably, BRAVEn-Large trained with around 3,000 hours of unlabelled data and only 30 hours of annotated data achieves 20.0\,\% / 1.7\,\% word error rate (WER) for VSR / ASR on the LRS3 test set, making it competitive with methods trained on orders of magnitude more transcribed data~\cite{makino2019recurrent,serdyuk2021audio}.

\label{sec:intro}

\begin{figure*}
  \centering
  \includegraphics[width=0.9\linewidth]{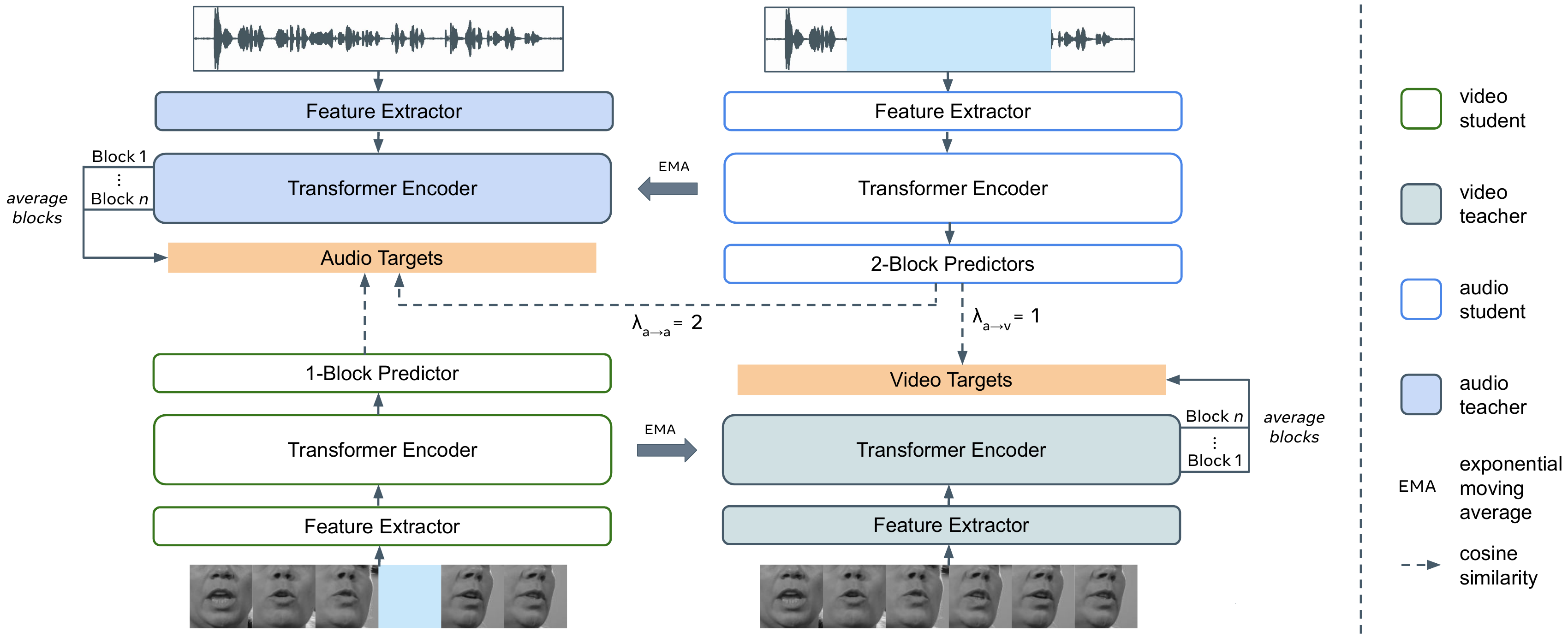}
  \caption{\textbf{BRAVEn overview}. BRAVEn uses targets that are the average of the outputs of the Transformer encoder blocks of the teacher networks. The video student predicts the audio targets via a 1-block predictor. In contrast, the audio student uses 2-block predictors and uses both a cross- as well as a within-modal loss, whose weight is twice as large as that of the cross-modal loss. The input masking for audio is stronger than the masking applied for video.}
  \label{fig:overview}
\end{figure*}
\section{Method: BRAVEn}
\subsection{Background: RAVEn}
RAVEn~\cite{haliassos2022jointly} learns visual and auditory speech representations by leveraging the correspondence between lip movements and the associated speech waveform, as well as employing masked prediction to encourage the networks to learn context-aware representations.

RAVEn uses a student and teacher for each modality.  The students intake masked video or audio and predict targets generated by the teachers, which are presented with unmasked inputs. Each student comprises a modality-specific convolutional feature extractor followed by two Transformer encoders: The first produces the desired representations, and the second predicts the targets given those representations in addition to mask tokens corresponding to the masked input. The teachers are architecturally identical to the students, but do not contain predictors. Denoting the student encoder weights for modality $m\in \{v,a\}$ as $s^{(m)}$ and the teacher weights as $t^{(m)}$, the following EMA update~\cite{grill2020bootstrap} is performed at each iteration:
\begin{align}
    t^{(m)}\leftarrow\mu t^{(m)}+(1-\mu)s^{(m)},
\end{align}
where $\mu$ follows a cosine schedule from 0.999 to 1.


The targets are simply the outputs of the teachers' Transformer encoders and are predicted by the students via cosine similarity. While the audio student predicts both the video and audio targets via separate predictors (corresponding to cross- and within-modal learning tasks, respectively), the video student predicts only the audio targets, due to the asymmetry in information content between the two modalities~\cite{haliassos2022jointly}. The losses for the video student, $\mathcal{L}_v$, and audio student,  $\mathcal{L}_a$, are given as
\begin{align}
    \mathcal{L}_v=\mathcal{L}^{v\to a}, \quad \mathcal{L}_a=\lambda_{a\to v}\mathcal{L}^{a\to v}+\lambda_{a\to a}\mathcal{L}^{a\to a} ,
\end{align}
where $\mathcal{L}^{v\to a}$, $\mathcal{L}^{a\to v}$, and $\mathcal{L}^{a\to a}$ denote the video-to-audio, audio-to-video, and audio-to-audio prediction losses, respectively, and $\lambda_{a\to v}$ and $\lambda_{a\to a}$ are scaling factors, which are both set to 1.
\subsection{BRAVEn}
In the following, we describe in detail each of BRAVEn's design improvements over RAVEn (see Figure~\ref{fig:overview}).

First, BRAVEn uses the mean of the outputs of all Transformer blocks rather than only the final block's output~\cite{lian2023av}, which is possible due to the feature size homogeneity across the Transformer architecture. This averaging operation likely results in smoother and higher-quality targets, which can aid the training dynamics~\cite{grill2020bootstrap}. Moreover, we apply instance normalisation~\cite{ulyanov2016instance} to the averaged targets to prevent representation collapse~\cite{caron2021emerging}. In RAVEn, the layer normalisation~\cite{ba2016layer} that follows the final block in the Transformer architecture serves a similar purpose.

Second, BRAVEn uses asymmetric predictor depths. RAVEn uses two-block Transformer predictors (with attention dimension 512), shown to work optimally \textit{when using the same design for both video and audio students}. BRAVEn instead uses a \textit{one-block} Transformer predictor for the video student (which predicts the audio targets), while retaining the original design of the audio predictors. The intuition is that audio is more relevant to speech recognition than video, and thus a shallower predictor leads to visual representations that more closely match the information in the audio targets.

Third, BRAVEn applies stronger masking to the audio inputs. Each corresponding video frame index has a 40\,\% probability of being picked as the start of a mask for audio, while a probability of 20\,\% is used for video\footnote{Then, the three consecutive video and 1,920 corresponding audio samples are zeroed out (since a video frame corresponds to 640 audio samples).}. In contrast, RAVEn uses a probability of 20\,\% for both modalities, which was chosen based on the assumption that \textit{both modalities would equally benefit from the same masking strategy}. BRAVEn's masking asymmetry is beneficial likely due to the difference in difficulty between VSR and ASR: Stronger masking for audio leads to more context-aware representations, but for video it may make the pretext task overly difficult.

Finally, BRAVEn uses asymmetric loss weights. In RAVEn, the weights of the two losses associated with the audio student are equal. BRAVEn instead uses a larger weight for the audio-to-audio ($\lambda_{a\to a}=2$) than the audio-to-video loss ($\lambda_{a\to v}=1$), which we empirically find leads to improvements for ASR.

\subsection{Fine-tuning}
After pre-training, all networks are discarded except for the video and audio student encoders, which are fine-tuned for VSR and ASR, following \cite{haliassos2022jointly}. We use the CTC / attention loss with a CTC weight of 0.1, which is the same weight used during decoding. We use a beam search size of 40, and our vocabulary is constructed using 1,000 SentencePiece~\cite{kudo2018sentencepiece} subword units. We also experiment with self-training: The fine-tuned ASR model is used to pseudo-label the unlabelled data, which is then used for fine-tuning the BRAVEn pre-trained encoders~\cite{shi2022learning,haliassos2022jointly,zhu2023vatlm}. We also present results when using the language model from \cite{ma2022visual,ma2023auto}, incorporated through shallow fusion. 


\section{Experimental Setup}
\subsection{Datasets}
We use the following datasets for pre-training: LRS3~\cite{afouras2018lrs3}, VoxCeleb2~\cite{DBLP:conf/interspeech/ChungNZ18}, and AVSpeech~\cite{ephrat2018looking}. LRS3 contains 433 hours of audio-visual speech. We use English-only versions of VoxCeleb2 and AVSpeech~\cite{shi2022learning,ma2022visual}, which comprise 1,326 and 1,323 hours of footage, respectively. We refer to a combination of LRS3 with VoxCeleb2 as LRS3+Vox2. When additionally using AVSpeech, we refer to the resulting dataset as LRS3+Vox2+AVS. For fine-tuning, we use either the 30-hour ``trainval'' set of LRS3 (``low-resource'' setting) or the full 433-hour LRS3 dataset (``high-resource'' setting).

\begin{table}
\centering
\resizebox{0.8\linewidth}{!}{
\begin{tabular}[b]{l c c c}\toprule
 & Base & Base+ & Large  \\ \midrule
Number of parameters (M) & 41 & 93 & 328 \\ 
Number of blocks & 12 & 12 & 24 \\
Attention dimension & 512 & 768 & 1024  \\ 
Number of attention heads & 8 & 12 & 16 \\
MLP size & 2048 & 3072 & 4096 \\ 
\bottomrule 
\end{tabular}}
\caption{\textbf{Configuration of the Transformer encoders.}}
\label{table:nets_config}
\end{table}

\subsection{Architecture}
Each encoder consists of a modality-specific, convolutional feature extractor followed by a Transformer encoder~\cite{vaswani2017attention,haliassos2022jointly}. The feature extractor for video is a 2D ResNet-18~\cite{he2016deep} with a 3D stem~\cite{stafylakis2017combining}, while for audio it is a 1D ResNet-18, as in \cite{haliassos2022jointly}. We use three different Transformer encoder sizes: Base, Base+, and Large. Details of their configuration are given in Table~\ref{table:nets_config}. We note that Base+ and Large are model sizes used in related works, such as \cite{shi2022learning} and \cite{lian2023av}. Although these works share the weights of the Transformer encoder for video and audio during pre-training, they separately fine-tune the models for VSR and ASR, resulting in separate sets of parameters.


For high-resource fine-tuning, we use Transformer decoders with 6 and 9 blocks for Base(+) and Large, respectively. The decoder dimensions (\textit{i.e.}, width) match the corresponding encoders. For low-resource, we use a decoder with 6 blocks and with hidden size, MLP size, and attention heads equal to 256, 2048, and 4, respectively, to avoid overfitting.

\subsection{Data pre-processing}
We follow the common protocol from \cite{ma2022visual,haliassos2022jointly,shi2022learning} for pre-processing. We crop a $96 \times 96$ area around the mouth and convert to grayscale. We use the original raw audio without any pre-processing. Any utterance longer than 24 seconds is divided into separate samples.

\subsection{Training details}
We use the AdamW~\cite{loshchilov2017decoupled} optimiser with a cosine decay learning scheduler. We pre-train the networks for 150 epochs with a warmup of 40 epochs for LRS3 and of 30 epochs when pre-training with more data. The learning rate is $3\times 10^{-3}$ when training BRAVEn-Base and Base+, and it is $2\times 10^{-3}$ and $1\times 10^{-3}$ when training BRAVEn-Large on LRS3+Vox2 and LRS3+Vox2+AVS, respectively. The weight decay is set to 0.04. We use drop path~\cite{huang2016deep} with a default value of 0.05, except when training the Large model on LRS3+Vox2, where it is set to 0.1. We use a maximum of 2,400 video frames per GPU for the Base and Base+ models and 900 frames per GPU for BRAVEn-Large. For video augmentations, we horizontally flip the video with 50\,\% probability and randomly crop them to size $88\times 88$. We do not augment the audio. For fine-tuning, we use the protocol from \cite{haliassos2022jointly}.




\begin{table*}[t]
\begin{subtable}[h]{0.48\linewidth}
\centering
\resizebox{\linewidth}{!}{
\begin{tabular}[b]{l c c c c c c}\toprule
\multirow{2}{*}{Method} & \multirow{2}{*}{Encoder} & \multirow{2}{*}{LM} & \multirow{2}{*}{Unlab hrs} & \multirow{2}{*}{Lab hrs} & \multicolumn{2}{c}{WER (\%)} \\
\cmidrule(lr){6-7}
& & & & & VSR & ASR \\
\midrule\midrule
\multicolumn{7}{c}{\textcolor{gray}{\textit{supervised (includes non-publicly available data)}}} \\
\textcolor{gray}{V2P~\cite{shillingford2018large}} & \textcolor{gray}{RNN} & \textcolor{gray}{\cmark} & \textcolor{gray}{-} & \textcolor{gray}{3,886} & \textcolor{gray}{55.1} & \textcolor{gray}{-} \\
\textcolor{gray}{RNN-T~\cite{makino2019recurrent}} & \textcolor{gray}{RNN} & \textcolor{gray}{\xmark} & \textcolor{gray}{-} & \textcolor{gray}{31,000} & \textcolor{gray}{33.6} & \textcolor{gray}{4.8} \\
\textcolor{gray}{VTP~\cite{afouras2021sub}} & \textcolor{gray}{Transf} & \textcolor{gray}{\cmark} & \textcolor{gray}{-} & \textcolor{gray}{2,676} & \textcolor{gray}{30.7} & \textcolor{gray}{-} \\ 
\textcolor{gray}{ViT3D-TM~\cite{serdyuk2021audio}} & \textcolor{gray}{Transf} & \textcolor{gray}{\xmark} & \textcolor{gray}{-} & \textcolor{gray}{90,000} & \textcolor{gray}{25.9} & \textcolor{gray}{2.3} \\
\textcolor{gray}{ViT3D-CM~\cite{DBLP:journals/corr/abs-2201-10439}} & \textcolor{gray}{Conf} & \textcolor{gray}{\xmark} & \textcolor{gray}{-} & \textcolor{gray}{90,000} & \textcolor{gray}{\textbf{17.0}} & \textcolor{gray}{\textbf{1.6}} \\
\midrule\midrule
\multicolumn{7}{c}{\textit{self-supervised}} \\
\textbf{Base(+) models} & & & & \\
AV-HuBERT~\cite{shi2022learning} & Transf & \xmark & 403 & 30 & 51.8 & 4.9 \\
VATLM\textsuperscript{$\dagger$}~\cite{zhu2023vatlm} & Transf & \xmark & 403 & 30 & 48.0 & - \\
RAVEn\textsuperscript{$\ddagger$}~\cite{haliassos2022jointly} & Transf & \xmark & 403 & 30 & 47.0 & 4.7 \\
AV-data2vec~\cite{lian2023av} & Transf & \xmark & 403 & 30 & 45.2 & 4.4 \\
\rowcolor{LightCyan}
BRAVEn\textsuperscript{$\ddagger$} & Transf & \xmark & 403 & 30 & \textbf{43.4} & \textbf{4.0} \\
\midrule
\textbf{Base(+) models} & & & & \\
AV-HuBERT~\cite{shi2022learning} & Transf & \xmark & 1,729 & 30 & 46.1 & 4.6 \\
VATLM\textsuperscript{$\dagger$}~\cite{zhu2023vatlm} & Transf & \xmark & 1,729 & 30 & 42.6 & - \\
RAVEn\textsuperscript{$\ddagger$}~\cite{haliassos2022jointly} & Transf & \xmark & 1,729 & 30 & 40.2 & 3.8 \\
AV-data2vec~\cite{lian2023av} & Transf & \xmark & 1,729 & 30 & 37.8 & 3.7 \\ 
\rowcolor{LightCyan}
BRAVEn & Transf & \xmark & 1,729 & 30 & \textbf{35.1} & \textbf{3.0} \\ 
\midrule
\textbf{Large models} & & & & \\
AV-HuBERT~\cite{shi2022learning} & Transf & \xmark & 1,729 & 30 & 32.5 & 2.9 \\
VATLM\textsuperscript{$\dagger$}~\cite{zhu2023vatlm} & Transf & \xmark & 1,729 & 30 & 31.6 & - \\
RAVEn~\cite{haliassos2022jointly} & Transf & \xmark & 1,729 & 30 & 32.5 & 2.7 \\
AV-data2vec~\cite{lian2023av} & Transf & \xmark & 1,729 & 30 & 30.8 & 2.7 \\
AV-HuBERT w/ ST~\cite{shi2022learning} & Transf & \xmark & 1,729 & 30 & 28.6 & - \\
VATLM\textsuperscript{$\dagger$} w/ ST~\cite{zhu2023vatlm} & Transf & \xmark & 1,729 & 30 & 27.6 & - \\
RAVEn w/ ST~\cite{haliassos2022jointly} & Transf & \xmark & 1,729 & 30 & 24.8 & 2.3 \\
\rowcolor{LightCyan}
BRAVEn & Transf & \xmark & 1,729 & 30 & 30.8 & 2.3 \\
\rowcolor{LightCyan}
BRAVEn & Transf & \xmark & 3,052 & 30 & 24.8 & 2.1 \\
\rowcolor{LightCyan}
BRAVEn w/ ST & Transf & \xmark & 3,052 & 30 & 21.3 & 1.9 \\
\rowcolor{LightCyan}
BRAVEn w/ ST & Transf & \cmark & 3,052 & 30 & \textbf{20.0} & \textbf{1.7} \\
\bottomrule 
\end{tabular}}
\caption{\textbf{Low-resource.}}
\label{table:low_resource}
\end{subtable}
\hfill
\begin{subtable}[h]{0.48\linewidth}
\centering
\resizebox{\linewidth}{!}{
\begin{tabular}[b]{l c c c c c c}\toprule
\multirow{2}{*}{Method} & \multirow{2}{*}{Encoder} & \multirow{2}{*}{LM} & \multirow{2}{*}{Unlab hrs} & \multirow{2}{*}{Lab hrs} & \multicolumn{2}{c}{WER (\%)} \\
\cmidrule(lr){6-7}
& & & & & VSR & ASR \\
\midrule\midrule
\multicolumn{7}{c}{\textcolor{gray}{\textit{semi-supervised (uses external models for pseudo-labelling)}}} \\
\textcolor{gray}{KD+CTC~\cite{afouras2020asr}} & \textcolor{gray}{CNN} & \textcolor{gray}{\cmark} & \textcolor{gray}{344} & \textcolor{gray}{433} & \textcolor{gray}{59.8} & \textcolor{gray}{-} \\
\textcolor{gray}{CM-aux~\cite{ma2022visual}} & \textcolor{gray}{Conf} & \textcolor{gray}{\cmark} & \textcolor{gray}{641} & \textcolor{gray}{818} & \textcolor{gray}{31.5} & \textcolor{gray}{-} \\
\textcolor{gray}{Auto-AVSR~\cite{ma2023auto}} & \textcolor{gray}{Conf} & \textcolor{gray}{\cmark} & \textcolor{gray}{2,630} & \textcolor{gray}{818} & \textcolor{gray}{19.1} & \textcolor{gray}{\textbf{1.0}} \\
\textcolor{gray}{SynthVSR Large~\cite{liu2023synthvsr}} & \textcolor{gray}{Conf} & \textcolor{gray}{\cmark} & \textcolor{gray}{2,630} & \textcolor{gray}{4,090\textsuperscript{*}} & \textcolor{gray}{\textbf{16.9}} & \textcolor{gray}{-} \\
\midrule\midrule
\multicolumn{7}{c}{\textit{self-supervised}} \\
\textbf{Base(+) models} & & & & & & \\
AV-HuBERT~\cite{shi2022learning} & Transf & \xmark & - & 433 & 44.0 & - \\
RAVEn\textsuperscript{$\ddagger$}~\cite{haliassos2022jointly} & Transf & \xmark & - & 433 & 39.1 & 2.2 \\ 
AV-data2vec~\cite{lian2023av} & Transf & \xmark & - & 433 & 39.0 & 2.0 \\ 
\rowcolor{LightCyan}
BRAVEn\textsuperscript{$\ddagger$} & Transf & \xmark & - & 433 & \textbf{36.0} & \textbf{1.9} \\ \midrule
\textbf{Base(+) models} & & & & & & \\
AV-HuBERT~\cite{shi2022learning} & Transf & \xmark & 1,326 & 433 & 34.8 & 2.0 \\
VATLM\textsuperscript{$\dagger$}~\cite{zhu2023vatlm} & Transf & \xmark & 1,326 & 433 & 34.2 & - \\
RAVEn\textsuperscript{$\ddagger$}~\cite{haliassos2022jointly} & Transf & \xmark & 1,326 & 433 & 33.1 & 1.9 \\ 
AV-data2vec~\cite{lian2023av} & Transf & \xmark & 1,326 & 433 & 32.9 & 1.7 \\
\rowcolor{LightCyan}
BRAVEn & Transf & \xmark & 1,326 & 433 & \textbf{28.8} & \textbf{1.4} \\
\midrule
\textbf{Large models} & & & & & & \\
AV-HuBERT~\cite{shi2022learning} & Transf & \xmark & 1,326 & 433 & 28.6 & 1.3 \\
VATLM\textsuperscript{$\dagger$}~\cite{zhu2023vatlm} & Transf & \xmark & 1,326 & 433 & 28.4 & - \\
RAVEn~\cite{haliassos2022jointly} & Transf & \xmark & 1,326 & 433 & 27.8 & 1.4 \\
AV-data2vec~\cite{lian2023av} & Transf & \xmark & 1,326 & 433 & 28.5 & 1.4 \\
u-HuBERT~\cite{hsu2022u} & Transf & \xmark & 1,326 & 433 & 27.2 & 1.4 \\
AV-HuBERT w/ ST~\cite{shi2022learning} & Transf & \xmark & 1,326 & 433 & 26.9 & - \\
VATLM\textsuperscript{$\dagger$} w/ ST~\cite{zhu2023vatlm} & Transf & \xmark & 1,326 & 433 & 26.2 & - \\
RAVEn w/ ST~\cite{haliassos2022jointly} & Transf & \xmark & 1,326 & 433 & 24.4 & 1.4 \\
\rowcolor{LightCyan}
BRAVEn & Transf & \xmark & 1,326 & 433 & 26.6 & 1.2 \\
\rowcolor{LightCyan}
BRAVEn & Transf & \xmark & 2,649 & 433 & 23.6 & 1.2 \\
\rowcolor{LightCyan}
BRAVEn w/ ST & Transf & \xmark & 2,649 & 433 & 20.9 & 1.2 \\
\rowcolor{LightCyan}
BRAVEn w/ ST & Transf & \cmark & 2,649 & 433 & \textbf{20.1} & \textbf{1.1} \\
\bottomrule 
\end{tabular}}
\caption{\textbf{High-resource.}}
\label{table:high_resource}
\end{subtable}
\caption{\textbf{LRS3 results}. ``Unlab hrs'' / ``lab hrs'' denote unlabelled hours / labelled hours. Self-supervised methods use labelled data for pre-training (without labels) along with unlabelled data (if any). ``LM'' denotes language model. ``ST'' denotes self-training. \textsuperscript{*}Includes synthetic visual data generated using labelled audio. \textsuperscript{$\dagger$}Uses an additional 3,846 / 452 hours of audio / audio-text data. \textsuperscript{$\ddagger$}Uses the (smaller) Base model.}
\end{table*}

\section{Results}
\subsection{Low-resource setting}
We provide results for the low-resource setting in Table~\ref{table:low_resource}. Using BRAVEn-Base and LRS3 as the pre-training dataset, we achieve better VSR and ASR word error rate (WER) than the previous state of the art obtained by AV-data2vec (43.4\,\% / 4.0\,\% vs 45.2\,\% / 4.4\,\% for VSR / ASR), despite using approximately half the number of parameters during inference. Using LRS3+Vox2 for pre-training, BRAVEn-Base+ obtains 35.1\,\% / 3.0\,\% WER for VSR / ASR, which represents the state of the art for this setting too. Increasing the model size to Large improves the results to 30.8\,\% and 2.3\,\% WER for VSR and ASR respectively, demonstrating BRAVEn's strong scalability. We are also the first to experiment with pre-training on LRS3+Vox2+AVS, consisting of 3,082 hours of unlabelled data. This increase in pre-training data results in significant WER improvement for VSR (30.8\,\%$\rightarrow$24.8\,\%) and modest improvement for ASR (2.3\,\%$\rightarrow$2.1\,\%). We believe we have not reached saturation yet, and increasing pre-training data further will yield even more improvements. Finally, using self-training and a language model during inference leads to 20.0\,\% / 1.7\,\% WER for VSR / ASR. We point out that our results with only \textit{30 hours} of labelled data (and no external pre-trained ASR model) are competitive with supervised methods that use orders of magnitude of more labelled data~\cite{makino2019recurrent,serdyuk2021audio}.

\subsection{High-resource setting}
The results for the high-resource setting are shown in Table~\ref{table:high_resource}. In each setting, BRAVEn achieves state-of-the-art performance among the self-supervised methods. Our best results for the high-resource setting are 20.1\,\% / 1.1\,\% for VSR / ASR. Interestingly, BRAVEn-Large coupled with self-training and a language model yields similar VSR results ($\sim$ 20\,\% WER) in both low- and high-resource settings, but better results are obtained in the high-resource setting for ASR. This suggests that high-quality transcriptions are more important for ASR than VSR. Furthermore, our results are competitive with semi-supervised methods that use external ASR models trained on tens of thousands of hours of audio data~\cite{ma2023auto, liu2023synthvsr}. All in all, our results suggest that BRAVEn can effectively translate unlabelled audio-visual data into strong speech recognition performance for both modalities.

\subsection{Audio-visual results}

\begin{table}
\centering
\resizebox{0.83\linewidth}{!}{
\begin{tabular}[b]{l l l l l}\toprule
 & & \multicolumn{3}{c}{SNR (dB)} \\ \cmidrule(lr){3-5}
 & \multicolumn{1}{c}{Clean} & \multicolumn{1}{c}{5} & \multicolumn{1}{c}{0} & \multicolumn{1}{c}{-5} \\ \midrule
AV-data2vec (AV) & 4.2 & - & - & - \\
RAVEn (AV) & 4.7 & 14.3 & 18.2 & 58.0  \\ \midrule
BRAVEn (A) & \textbf{4.0} & 15.6 & 24.6 & 99.0  \\ 
BRAVEn (AV) & \textbf{4.0} \textcolor{Gray}{$\scriptstyle = $} & \textbf{12.4} \textcolor{ForestGreen}{$\scriptstyle \downarrow 3.2$} & \textbf{15.0} \textcolor{ForestGreen}{$\scriptstyle \downarrow 9.6$} & \textbf{48.5} \textcolor{ForestGreen}{$\scriptstyle \downarrow 50.5$}  \\
\bottomrule 
\end{tabular}}
\caption{\textbf{Audio-visual results.} We show low-resource results (WER~\%) (Base models) with different signal-to-noise ratio (SNR).}
\label{table:av_results}
\end{table}


Although not the focus of this work, we now provide results for audio-visual speech recognition using the late-fusion method proposed in \cite{ma2021end}. To that end, we append a fusion module that concatenates the features from the video and audio encoders and applies a two-layer MLP with hidden dimension 1024. We transfer the weights from the VSR and ASR encoders and then train the fusion model, CTC layer, and decoder for 200 epochs, keeping the encoder weights frozen. 

Results for the low-resource setting with the Base model and different levels of audio babble noise from the NOISEX corpus~\cite{varga1993assessment} are given in Table~\ref{table:av_results}. It is evident that as the audio noise level increases, so does the difference in performance between the audio-only and audio-visual models. BRAVEn outperforms RAVEn at all noise levels and achieves better WER than AV-data2vec, \textit{even when BRAVEn uses only the audio modality}.

\subsection{Ablations}
\begin{table}
\centering
\resizebox{0.85\linewidth}{!}{
\begin{tabular}[b]{l l l}\toprule
 & \multicolumn{2}{c}{WER (\%)} \\  \cmidrule(lr){2-3}
& VSR & ASR \\ \midrule
RAVEn & 47.0 & 4.7  \\
\hspace{2mm} + average blocks & 45.3 \textcolor{ForestGreen}{$\scriptstyle \downarrow 1.7$} & 4.6 \textcolor{ForestGreen} {$\scriptstyle \downarrow 0.1$} \\
\hspace{2mm} + shallower video predictor & 44.1 \textcolor{ForestGreen}{$\scriptstyle \downarrow 1.2$} & 4.6 \textcolor{Gray} {$\scriptstyle = $} \\
\hspace{2mm} + stronger audio masking & 43.5 \textcolor{ForestGreen}{$\scriptstyle \downarrow 0.6$} & 4.2 \textcolor{ForestGreen}{$\scriptstyle \downarrow 0.4$} \\
\hspace{2mm} + different loss weights = \textbf{BRAVEn} & \textbf{43.4} \textcolor{ForestGreen}{$\scriptstyle \downarrow 0.1$} & \textbf{4.0} \textcolor{ForestGreen}{$\scriptstyle \downarrow 0.2$} \\ \midrule
BRAVEn + audio mask probability 0.6 & 44.4 \textcolor{red}{$\scriptstyle \uparrow 1.0$} & 4.2 \textcolor{red}{$\scriptstyle \uparrow 0.2$} \\
BRAVEn + averaging last 6 blocks & 46.2 \textcolor{red}{$\scriptstyle \uparrow 2.8$} & 4.1 \textcolor{red}{$\scriptstyle \uparrow 0.1$} \\
\bottomrule 
\end{tabular}}
\caption{\textbf{Ablations.} We show results for the low-resource setting using the Base model.}
\label{table:ablations}
\end{table}

We conduct an ablation study to show the effect of each design choice on the VSR and ASR performance (see Table~\ref{table:ablations}). Each new addition improves the VSR and / or ASR performance. We observe that averaging targets and using a shallower video predictor has a larger effect on VSR than ASR, while stronger audio masking and different loss weights have a more significant impact on ASR. Furthermore, using even stronger audio masking or using the average of the last 6 Transformer blocks as targets lead to worse performance.
\section{Conclusion}
We have presented BRAVEn, a self-supervised framework for learning visual and auditory speech representations from audio-visual data. We propose some design improvements to the recent RAVEn method, which cumulatively have a significant impact on the downstream VSR and ASR performance and achieve state-of-the-art results for audio-visual self-supervised methods in various settings. Furthermore, we experiment with doubling the amount of unlabelled data during pre-training as used in other self-supervised works and observe strong scaling behaviour, even when fine-tuning with only 30 hours of labelled data. Our work provides compelling evidence that readily accessible unlabelled audio-visual data can effectively substitute costly annotated samples.

\vfill\pagebreak

\clearpage
\centering\section*{References}
\label{sec:refs}
\AtNextBibliography{\ninept}
\printbibliography[heading=none]

\end{document}